\documentclass[lettersize,journal]{IEEEtran}
\usepackage{amsmath,amsfonts}
\usepackage{algorithmic}
\usepackage{algorithm}
\usepackage{array}
\usepackage{subcaption}
\usepackage{textcomp}
\usepackage{stfloats}
\usepackage{url}
\usepackage{verbatim}
\usepackage{graphicx}
\usepackage{cite}
\usepackage{multirow}
\usepackage[hidelinks]{hyperref}
\usepackage{cleveref}
\crefname{section}{Section}{Sections}
\crefname{table}{Table}{Tables}
\crefname{figure}{Fig.}{Figs.}

\usepackage{tikz,xcolor,hyperref}
\definecolor{lime}{HTML}{A6CE39}
\DeclareRobustCommand{\orcidicon}{%
    \begin{tikzpicture}
    \draw[lime, fill=lime] (0,0) 
    circle [radius=0.16] 
    node[white] {{\fontfamily{qag}\selectfont \tiny ID}};    \draw[white, fill=white] (-0.0625,0.095) 
    circle [radius=0.007];    \end{tikzpicture}
    \hspace{-2mm}}
\foreach \x in {A, ..., Z}{%
    \expandafter\xdef\csname orcid\x\endcsname{\noexpand\href{https://orcid.org/\csname orcidauthor\x\endcsname}{\noexpand\orcidicon}}
    }
    
\begin{document}

\title{Temporal-Spatial Object Relations Modeling for Vision-and-Language Navigation}

\author{Bowen Huang\orcidA{},
Yanwei Zheng\IEEEauthorrefmark{1} \orcidB{}, \IEEEmembership{Member, ~IEEE},
Chuanlin Lan\orcidC{},
Xinpeng Zhao\orcidD{},
Yifei Zou\orcidE, \IEEEmembership{Member, ~IEEE},
Dongxiao yu\orcidF{}, \IEEEmembership{Senior Member, ~IEEE}
\thanks{
\IEEEauthorrefmark{1}Corresponding author.

Bowen Huang, Yanwei Zheng, Xinpeng Zhao, Yifei Zou, and Dongxiao Yu are with the School
of Computer Science and Technology, Institute of Intelligent Computing,
Shandong University, Qingdao 266237, China (e-mail: huangbw@mail.sdu.edu.cn; zhengyw@sdu.edu.cn; zhaoxp1001@gmail.com; yfzou@sdu.edu.cn, dxyu@sdu.edu.cn).
 
Chuanlin Lan is a Ph.D. student in the Department of Electrical Engineering at City University of Hong Kong, Hong Kong 999077, China(e-mail: cllan2-c@my.cityu.edu.hk).
}
}
\maketitle

\begin{abstract}
Vision-and-Language Navigation (VLN) is a challenging task where an agent is required to navigate to a natural language described location via vision observations.
The navigation abilities of the agent can be enhanced by the relations between objects, which are usually learned using internal objects or external datasets.
The relationships between internal objects are modeled employing graph convolutional network (GCN) in traditional studies.
However, GCN tends to be shallow, limiting its modeling ability.
To address this issue, we utilize a cross attention mechanism to learn the connections between objects over a trajectory, which takes temporal continuity into account, termed as Temporal Object Relations (TOR).
The external datasets have a gap with the navigation environment, leading to inaccurate modeling of relations.
To avoid this problem, we construct object connections based on observations from all viewpoints in the navigational environment, which ensures complete spatial coverage and eliminates the gap, called Spatial Object Relations (SOR).
Additionally, we observe that agents may repeatedly visit the same location during navigation, significantly hindering their performance. 
For resolving this matter, we introduce the Turning Back Penalty (TBP) loss function, which penalizes the agent's repetitive visiting behavior, substantially reducing the navigational distance.
Experimental results on the REVERIE, SOON, and R2R datasets demonstrate the effectiveness of the proposed method.
\end{abstract}


\section{Introduction}
\IEEEPARstart{I}{n} recent years, advancements in vision-and-language navigation (VLN) have demonstrated significant potential to revolutionize intelligent transportation systems, by enabling more intuitive and effective ways for autonomous vehicles to interpret and navigate complex urban environments~\cite{SFT, ADP}.
The goal of VLN~\cite{AirBERT, R2R, DIAG, REVERIE} is to guide an agent to a designated location, which is described by a natural language instruction. 
Although related vision-and-language problems have been extensively studied~\cite{IC, VQA, VisDial, RE}, the VLN task remains highly challenging due to the dynamic nature of real-world environments and the complexity of the instructions involved.

Significant achievements have been made in the field of VLN~\cite{R2R, VLNBERT}. 
Most previous methods~\cite{speaker, SOTA, BTED, ADAPT} employ RNN models~(\textit{e.g.}, GRUs and LSTMs) or transformer-based architectures to receive and store visual information, aligning it with the instruction to predict actions at each step.
In recent developments, a number of techniques~\cite{DUET, KERM, GridMM} have incorporated topological maps and semantic graphs to retain historical data, showing promising outcomes.
Concurrently, a subset of research~\cite{ORG, DOA, DAT} has started to underscore the importance of modeling relationships between objects in navigational environments. 

\begin{figure*}[!htbp]
\centering
  \begin{tabular}[b]{cc}
    \begin{tabular}[b]{c}
      \begin{subfigure}[b]{0.32\linewidth}
        \includegraphics[width=\textwidth]{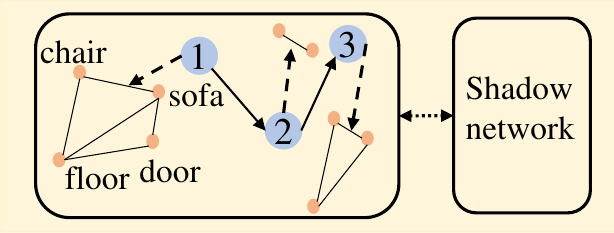}
        \caption{Graph-structured features}
        \label{fig:A}
      \end{subfigure}\\
      \begin{subfigure}[b]{0.32\linewidth}
        \includegraphics[width=\textwidth]{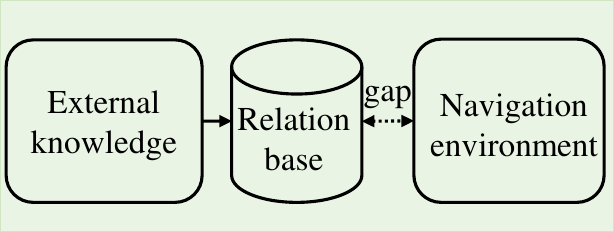}
        \caption{External knowledge}
        \label{fig:B}
      \end{subfigure}
    \end{tabular}
    &
    \begin{subfigure}[b]{0.6\linewidth}
      \includegraphics[width=\textwidth]{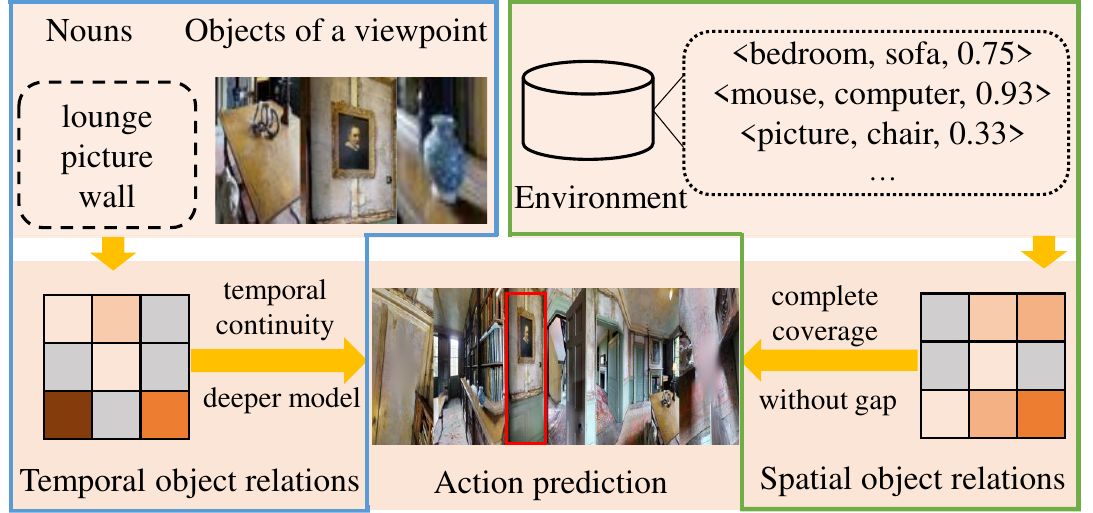}
      \caption{Temporal-spatial object relations modeling}
      \label{fig:C}
    \end{subfigure}
  \end{tabular}
  \label{fig:ABC}
  \caption{Three methods of learning the connections between objects}
\end{figure*}

A method for learning the relations between objects is constructing graph-structured feature representations~\cite{ORG, SEvol}.
As depicted in~\cref{fig:A}, a graph-based navigation state is maintained utilizing GCN at each location during the agent's navigation process (such as moving from position 1 to position 3). 
However, due to the issue of over-smoothing, GCN networks are typically kept shallow, which can impede their ability to accurately learn relationships.
Another approach of modeling the object relations is introducing external knowledge.
The external knowledge primarily originates from three sources: it is either derived from publicly available image-text datasets~\cite{BPOD, VG, KERM}; acquired through the use of the pretrained ConceptNet system~\cite{Conceptnet, CKR}; or obtained by employing large language models~\cite{VELMA, NavGPT, LM-Nav}.
Nevertheless, as shown in~\cref{fig:B}, this external knowledge does not directly come from within the navigational environment, resulting in a significant gap between the object relations it forms and those in the real environment.

To address the above problems, we present two fundamental modules: the Temporal Object Relations (TOR) module and the Spatial Object Relations (SOR) module.
As illustrated in~\cref{fig:C}, the TOR module captures the variations in positional relationships among objects from the moving trajectory of the agent.
At each position along its trajectory, a cross attention mechanism is used to calculate a relationship matrix based on objects found and nouns in the instruction. 
This matrix is consistently updated as the agent moves, thereby modeling the object relations from a continuous temporal perspective.
On the other hand, the SOR module is designed to model the spatial relations among objects from all locations within a navigational environment.
Each viewpoint is considered equally in the environment, creating a relational database based on the objects discovered at these locations. 
This database includes the information of all viewpoints and ensures complete spatial coverage.

While these modules enhance the agent's ability to understand and navigate complex environments, they also introduce a new challenge. 
The detailed tracking and continuous updating of object relations can lead the agent to explore new locations that are not part of the correct path or to conduct multiple explorations at the same location. 
Exploration across new locations is helpful, granting agents critical environmental insights and informing their navigational decisions~\cite{DUET, perception}. 
However, repetitive revisits to the same viewpoint do not enhance navigational success but rather impair efficiency.
To counteract this, we introduce the Turning Back Penalty (TBP) loss function. 
Specifically, during the training process of the agent, it penalizes the agent each time it passes a previously visited location. 
This effectively mitigates the issue of revisits, thereby improving navigation efficiency.
Our primary contributions can be summarized as follows:
\begin{itemize}
    \item[$\bullet$] We propose the TOR and SOR modules, which learn the interdependent relations among different objects from the dimensions of time and space, respectively.
    \item[$\bullet$] We introduce the TBP loss function, which effectively alleviates the problem of excessive path length caused by repeated visits to the same location by the agent.
	\item[$\bullet$] Extensive experiments have been conducted on the REVERIE~\cite{REVERIE}, SOON~\cite{SOON}, and R2R~\cite{R2R} datasets to demonstrate the superiority of our method over existing approaches in visual-and-language navigation.
\end{itemize}

The rest of the paper is organized as follows.
\cref{related rork} reviews relevant research about VLN.
\cref{method} introduces the details of our method.
In~\cref{experiment}, we present the training methodology and parameter settings of our model, and evaluate it on three datasets.
We conclude the paper in~\cref{conclusion}.

\section{Related work}\label{related rork}
\subsection{Vision-and-Language Navigation}
VLN~\cite{Reinforced, LVER, SSM, HOP, LU} has received significant research interests in recent years with the continual improvement.
Early methods~\cite{speaker, RCM, Reinforced} usually utilize recurrent neural networks (RNNs) to encode historical observations and actions, which are represented as a state vector. 
In order to capture environment layouts, Wang \textit{et al.}~\cite{SSM} employs a structured scene memory to accurately memorize the percepts during navigation.
Tan \textit{et al.}~\cite{BTED} proposes a two-stage training approach to enhance the generalization ability of agent.
Ma \textit{et al.}~\cite{Regretful} use a progress monitor as a learnable heuristic for search.

More recently, transformer-based architectures have been shown successful in VLN tasks~\cite{pretrain3}, notably by leveraging pre-trained architectures.
PRESS~\cite{PRESS} proposes a stochastic sampling scheme to reduce the considerable gap between the expert actions in training and sampled actions in test. 
VLN-BERT~\cite{VLNBERT} employs recurrent units in transformer architecture to predict actions.
To learn general navigation oriented textual representations, both AirBERT~\cite{AirBERT} and HM3D-AutoVLN~\cite{LU} introduce expansive VLN dataset to enhance the interaction between various modalities.
DUET~\cite{DUET} adeptly merges local observations with the overarching topological map through the use of graph transformers. 
This streamlines action planning and bolsters cross-modal comprehension.
GridMM~\cite{GridMM} builds a top-down egocentric and dynamically growing grid memory map to structure the visited environment.
Different from them, our work propose an object-relations model designed to enhance the agent’s understanding of its environment. 

\subsection{Object Relations Modeling}
Recently, some studies have begun to focus on utilizing the relationships between objects to guide agent navigation~\cite{ORG, HOZ}.
ORG~\cite{ORG} improves visual representation learning by integrating object relationships, including category closeness and spatial correlations.
SEvol~\cite{SEvol} propose a novel structured state-evolution model to learn the object-level relationship.
CKR~\cite{CKR} propose a knowledge-enabled entity relationship reasoning module to learn the internal-external correlations among room- and object-entities.
KERM~\cite{KERM} constructs an external knowledge base to assist in establishing relationships between the various entities described in the instructions.
OAAM~\cite{OA} utilizes two learnable attention modules to highlight language relating to objects and actions within a given instruction.
VLMaps~\cite{VLMaps} translates natural language commands into a sequence of open-vocabulary navigation goals using large language models, resulting in objects that are spatially defined.
Our method differs from others in modeling object relationships.
It both considers temporal continuity and ensures complete spatial coverage.
This integration results in a more accurate representation of the connections between objects.

\subsection{Training Regimes}
Previous studies~\cite{DUET} mostly train the agent with the supervision from a pseudo interactive demonstrator similar to the DAgger algorithm~\cite{reduction}.
Anderson \textit{et al.}~\cite{R2R} introduces two distinct training regimes, teacher-forcing and student-forcing, and utilizes cross entropy loss at each step to maximize the likelihood of the ground-truth target action.
Ma \textit{et al.}~\cite{SMNA} introduces a self-monitoring loss function that enhances the agent's performance by estimating its navigation progress.
Tan \textit{et al.}~\cite{BTED} introduce an environment dropout method and enable the agent to navigate in environments with incomplete information and improving its generalization.
Wang \textit{et al.}~\cite{Reinforced} introduces a loss function that integrates reinforcement learning and self-supervised learning to optimize the agent's matching capability across different modalities.
These methods accumulate penalties for the agent as the number of exploratory steps increases.
However, due to the poor balancing of penalty intensity, this may lead to either excessive revisitation of the same location by the agent or insufficient exploration of the environment. 
In contrast, our TBP loss function avoids such pitfalls by preventing the agent from retracing its steps while ensuring ample exploration of the environment.

\begin{figure*}[!htbp]
	\centering
	\includegraphics[width=\linewidth]{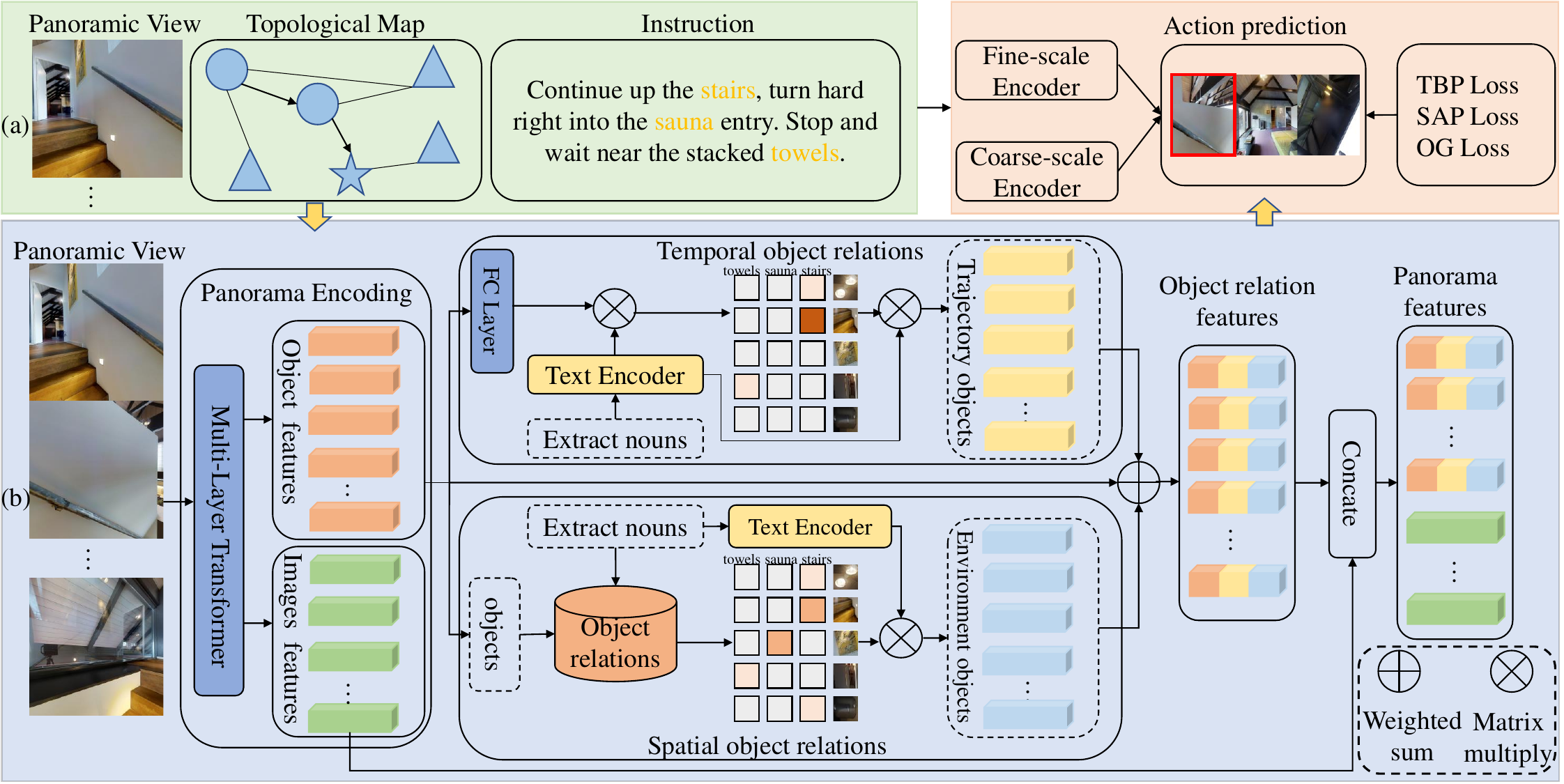}
	
	\caption{The overall network architecture. (a) The baseline utilizes a dual-scale encoder to encode local panoramic features, global historical features, and instruction features for action prediction of the agent. (b) At each time step $t$, our method employ two modules to learn temporal and spatial object relations. Then the object relation features are combined with the image features for action prediction. Finally, we designs a novel TBP loss function to supervise the training of the agent in order to reduce its tendency to backtrack.}
 
	\label{frameworkl}
\end{figure*}

\section{Method}\label{method}
In the VLN task, the agent is initially located at a starting node in a previously unseen environment. 
The environment is represented by a weighted undirected graph $\mathcal{G}=\left \{\mathcal{V},\mathcal{E}\right \}$, where $\mathcal{V}$ denotes navigable nodes and $\mathcal{E}$ denotes edges. 
The agent needs to explore this environment to reach the target location, guided by a natural language instruction.
The instruction embedding consisting of $L$ words is $\mathcal{W}=\left\{\boldsymbol{w}_i\right\}_{i=1}^{L}$.
At each time step $t$, the agent observes a panoramic view of the current node.
The panorama is divided into $n$ different perspective images $\mathcal{R}_t=\left\{\boldsymbol{r}_i\right\}_{i=1}^n$, where $\boldsymbol{r}_i$ denotes the image feature of the i-th perspective and the direction encoding of that perspective. 
In addition, $m$ object features $\mathcal{O}_t=\left\{\boldsymbol{o}_i\right\}_{i=1}^m$ are extracted from the panorama. 
This is done using annotated object bounding boxes or automatic object detectors~\cite{BUTD}, enhancing the agent's fine-grained visual perception.

\subsection{Overview of Our Approach}
As shown in~\cref{frameworkl}(a), we adopt the architecture of DUET~\cite{DUET} as the baseline, It consists of three inputs: panoramic visual features of the current location, a topological map, and an instruction.
At time step $t$, the topological map is represented as $\mathcal{G}_t=\left \{\mathcal{V}_t,\mathcal{E}_t\right \}$.
Here, $\mathcal{G}_t$ is a subset of the overall map $\mathcal{G}$ and encapsulates the state of the environment after $t$ steps of navigation. 
$\mathcal{V}_t$ contains three kinds of nodes: visited nodes (circular nodes), the current node (pentagram nodes), and navigable nodes (triangular nodes).

Our method, as illustrated in~\cref{frameworkl}(b), employs two modules to calculate temporal object features $\mathcal{M}_t$ and spatial object features $\mathcal{N}_t$ at each step $t$, respectively. 
These features are then combined with $\mathcal{O}_t$ and $\mathcal{R}_t$, in order to generate the panoramic feature $\mathcal{Q}_t$, which is fed into a dual-scale encoder to predict the agent's action. 
To further enhance the agent's performance, we introduce the TBP loss function.
It can help to prevent repeated explorations and reduce the length of the agent's path.

\subsection{Object Relations}
In the VLN task, there are connections between the various objects that the agent perceives. 
In our method, these connections are learned from two dimensions of time and space, significantly enhancing the accuracy of the navigation.

\subsubsection{Object nouns features}
To extract object-related noun features from the instruction, we begin by choosing word embeddings describing objects from $\mathcal{W}$. 
Specifically, our process begins by obtaining labels for all objects from the MatterPort3D simulator~\cite{MP3D}. 
These labels are then compiled into a noun database, denoted as $D$. 
For a given natural language instruction $\mathcal{W}$, we iterate through each word. 
If a word is found within $D$, it is selected as an object-related token.

Upon acquiring these object-related embeddings, we enhance them with positional embeddings as described in~\cite{BERT}. 
These positional embeddings correspond to the respective word's location within the sentence. 
Additionally, a type embedding specific to text, as outlined in~\cite{LXMERT}, is also incorporated.
Next, we input all noun tokens into a text encoder that consists of a multi-layer transformer. 
This process generates contextual noun representations, which we refer to as ${\mathcal{\hat{W}}} = \{\boldsymbol{\hat{w}}_1, \boldsymbol{\hat{w}}_2, \dots, \boldsymbol{\hat{w}}_{\hat{L}}\}$.

\begin{figure*}[!htbp]
    \centering
    \begin{subfigure}[t]{0.48\textwidth}
        \centering
        \includegraphics[width=\linewidth]{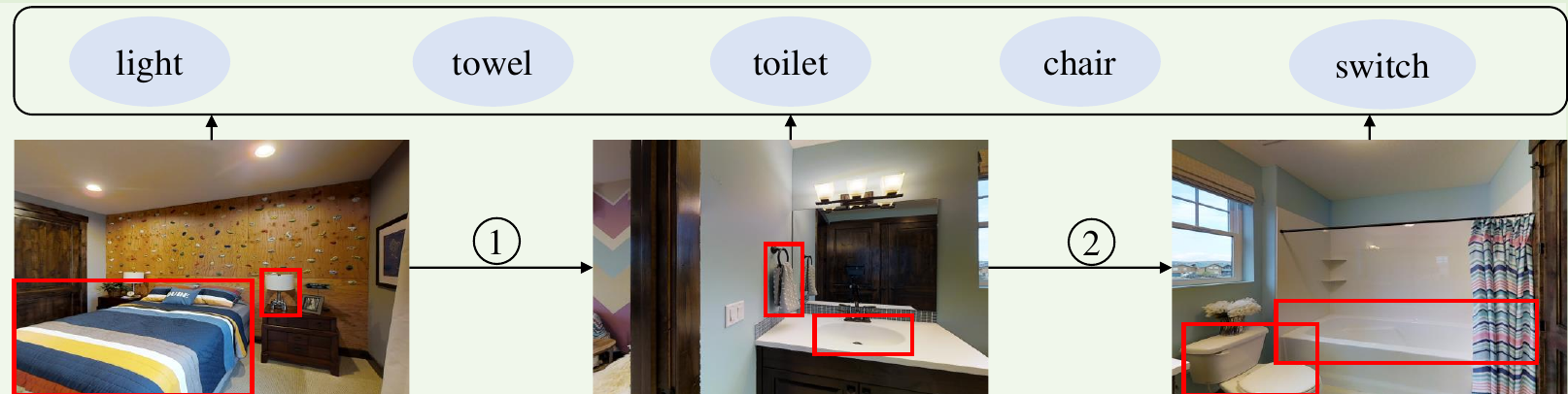} 
        \caption{As the agent interacts with the environment, it continuously updates its understanding of these relationships.}
        \label{relation1}
    \end{subfigure}
    \hfill
    \begin{subfigure}[t]{0.48\textwidth}
        \centering
        \includegraphics[width=\linewidth]{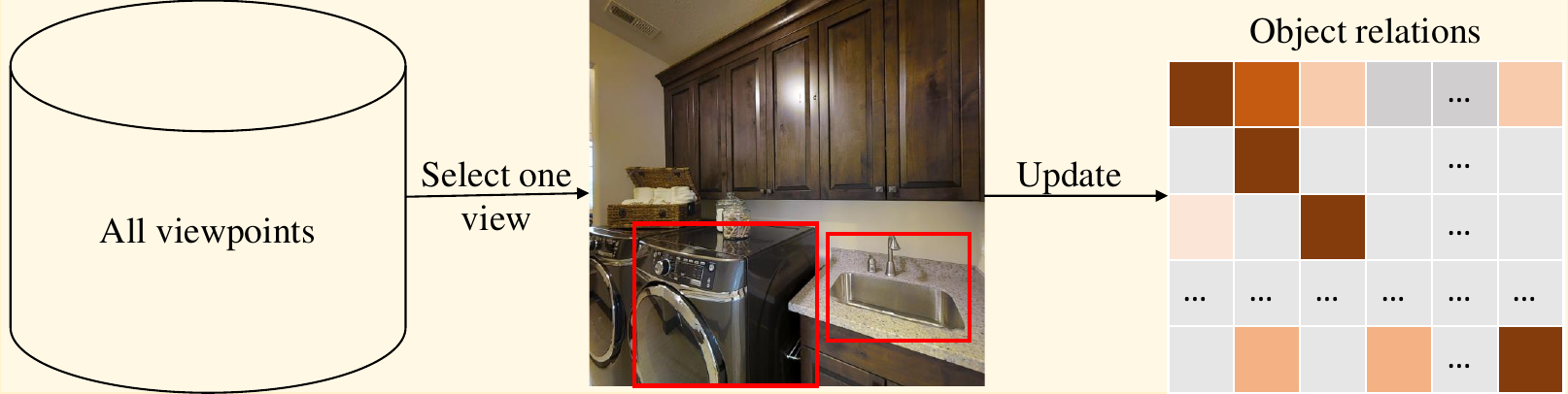}
        \caption{The spatial object relationship matrix is constructed by considering the objects visible at each node in the training environment.}
        \label{relation2}
    \end{subfigure}%
    \caption{Learning methods for two kinds of relationships}
    \label{framework}
\end{figure*}

\subsubsection{Temporal object relations}
In the study of object relations, to circumvent the issue of inadequate learning capabilities resulting from the shallow nature of GCN, we have designed a temporal object relations (TOR) module.
This module employs a cross attention mechanism to focus on objects observed during navigation and noun embeddings in the instruction.
Through this approach, we learn a relationship matrix accurately as the agent progresses along its exploration trajectory.
This ensures temporal continuity in the agent's learning process.

\cref{relation1} depicts how the agent, when arriving at a new location, establishes connections between the objects it perceives and the relevant nouns from the navigation instruction.
Specifically, when the agent reaches a location, it obtains a panoramic view of that position. 
The agent employs a cross attention mechanism to learn the associations between all objects discovered at this location and the nouns mentioned in the instruction. 
This process is consistently applied at each location along the agent's navigational path.
It sets the stage for the agent to learn and progressively refine the inter-object relationships across temporal dimensions.

In our approach, we treat the object features, denoted as $O_t$, as the query, and the noun features, $\hat{W}$, as the key. 
We employ a cross attention mechanism to compute the relationship matrix $T$, which can be expressed in the following way:
\begin{equation}
\mathbf{T} = {\rm FC}\left(\mathcal{O}_t\right) \mathcal{\hat{W}}
\end{equation}
where $\rm FC$ is a fully connected layer.

We leverage $\mathbf{T}$ to derive the temporal object features $\mathcal{M}_t$.
It is formalized as follow:
\begin{equation}
\mathcal{M}_t = \mathbf{T} \mathcal{\hat{W}}
\end{equation}

\subsubsection{Spatial object relations}
Due to the gap between external knowledge and the navigation environment, the agent is unable to accurately learn the relationships between objects based on external knowledge. 
To bridge this gap, we introduce the spatial object relations (SOR) module.
This module considers all viewpoints in the environment and ensures complete spatial coverage, which help the agent learn a more precise object relations.

During the establishment of spatial object relations, the agent proceeds to update an initially zero-valued relationship matrix $\mathbf{E}$ based on the objects identified at each respective location. 
As illustrated in~\cref{relation2}, for objects $x$ and $y$ observed concurrently at the same location, a shorter distance between them correlates with a stronger association. 
Consequently, we update the relationship matrix based on the distances between objects as seen by the agent at each location within the environment.
The update rule for $\mathbf{E}$ is formalized as follows:
\begin{equation}
\mathbf{E}\left(x, y\right) += \frac{k_1}{k_2 \Vert \boldsymbol{v}_x - \boldsymbol{v}_y \Vert_2 + k_3 \Vert \boldsymbol{d}_x - \boldsymbol{d}_y \Vert_2}
\end{equation}
where $\boldsymbol{v}_x$ and $\boldsymbol{v}_y$ denote the perspectives from the current position in relation to objects $x$ and $y$, respectively, and $\boldsymbol{d}_x$ and $\boldsymbol{d}_y$ represent the respective depths of objects $x$ and $y$ from the current position. 
Here, $k_1$, $k_2$, and $k_3$ are predefined constants, and $\Vert \cdot \Vert_2$ denotes the L2 norm.

In the course of the agent's training, the pertinent matrix $\mathbf{E}^{\prime}$ is retrieved from $\mathbf{E}$, guided by the objects detected at the agent’s current location and the nouns encapsulated within the instruction. 
Assuming the agent currently discovers $c_1$ objects, and the current instruction contains $c_2$ nouns, the calculation of $\mathbf{E}^{\prime}$ is as follows:

\begin{equation}
\resizebox{0.9\hsize}{!}{
$\mathbf{E}^{\prime} = \begin{bmatrix}
    \mathbf{E}_{p(1)q(1)} & \mathbf{E}_{p(1)q(2)} & \mathbf{E}_{p(1)q(3)} & \dots & \mathbf{E}_{p(1)q(c_2)} \\
    \mathbf{E}_{p(2)q(1)} & \mathbf{E}_{p(2)q(2)} & \mathbf{E}_{p(2)q(3)} & \dots & \mathbf{E}_{p(2)q(c_2)}\\
    \dots & \dots & \dots & \dots & \dots\\
    \mathbf{E}_{p(c_1)q(1)} & \mathbf{E}_{p(c_1)q(2)} & \mathbf{E}_{p(c_1)q(3)} & \dots & \mathbf{E}_{p(c_1)q(c_2)}\\
\end{bmatrix}$}
\end{equation}
where $p(i)$ is the index of the i-th object and $q(j)$ is the index of the j-th noun.
Subsequently, matrix multiplication is employed to derive the environmental object feature $\mathcal{N}_t$, as per the following equation:
\begin{equation}
\mathcal{N}_t = \mathbf{E}^{\prime} \mathcal{\hat{W}}
\end{equation}

Upon obtaining both the temporal object features $\mathcal{M}_t$ and the spatial object features $\mathcal{N}_t$, the final object relationship feature $\mathcal{Q}_t$ is computed utilizing the equation:
\begin{equation}
\mathcal{Q}_t = \alpha_1 \mathcal{O}_t + \alpha_2 \mathcal{M}_t + \alpha_3 \mathcal{N}_t
\end{equation}
where $\alpha_1$, $\alpha_2$, and $\alpha_3$ are three learnable parameters, summing to 1. 
Subsequently, a concatenation of $\mathcal{Q}_t$ and the image feature $\mathcal{R}_t$ is performed to yield the panoramic feature $\mathcal{F}_t = \left[\mathcal{R}_t, \mathcal{Q}_t\right]$

\subsection{Turning back penalty} 
After introducing the TOR and SOR modules, we observed the agent tended to revisit the same location multiple times.
This usually culminates in an elongated navigation path. 
For instance, an agent's journey initiates at point $a$ and concludes at point $e$, successfully navigating through the route $a \rightarrow b \rightarrow c \rightarrow d \rightarrow b \rightarrow e$. 
It can be observed that the direct path composed of $a \rightarrow b \rightarrow e$ would be optimal. 
The exploration of vertices $c$ and $d$ represents additional exploration by the agent, while revisiting vertex $b$ indicates redundant exploration. 
Such additional exploration is beneficial as it enables the agent to acquire new knowledge, thereby enhancing its navigational skills. 
Conversely, redundant exploration does not improve navigation efficiency and leads to unnecessarily prolonged paths, which is disadvantageous.

To address the issue of the agent frequently revisiting the same location, we have developed a new loss function, named Turning Back Penalty (TBP).
This function introduces a punitive measure to discourage the agent from redundant navigation, fostering a more streamlined and direct trajectory.
Concretely, let us consider a scenario wherein the agent is positioned at location $a$, and it has a set of $r$ navigable positions, denoted as $\left\{b_1, b_2, \dots, b_r\right\}$. 
\begin{equation}
L_{TBP} = \sum_{i=1}^{r}\frac{e^{p_i}d_i}{\sum_{j=1}^{r}e^{p_j}}
\end{equation}
where $p_i$ symbolizes the probability of transitioning from location $a$ to location $b_i$, and $d_i$ represents the cumulative length of the paths that have been traversed repetitively by the agent in the course of navigating from $a$ to $b_i$.

\subsection{Training and Inference}
\subsubsection{Pretrainging}
Building upon the foundational work presented in~\cite{pretrain1, pretrain2, pretrain3}, the advantageous impacts of pretraining transformer-based models within the domain of Vision-and-Language Navigation (VLN) have been well-established.
In this work, we employ four auxiliary tasks to pretrain our proposed model.
\paragraph{Masked Language Modeling (MLM)}
In accordance with~\cite{BERT}, we randomly mask out certain words in the instructions with a probability of 15\%.
Subsequently, we use the contextual information within the instructions to predict the masked words.

\paragraph{Masked Region Classification (MRC)}
In the MRC~\cite{MRC} task, the model need to identify the semantic labels of images where certain views are masked. 
We employ a random masking of view images at a rate of 15\%. 
In alignment with the methodology in~\cite{DUET}, we use an image classification model~\cite{ViT} pre-trained on ImageNet to generate target labels for these view images.

\paragraph{Single-step Action Prediction (SAP)}
In the SAP~\cite{SAP} task, the agent is required to infer its next action based on its previous actions.
The SAP loss in behavior cloning given a demonstration path $P^*$ is as follows:
\begin{equation}
    L_{SAP} = \sum_{t=1}^{T}-\log p\left(a_t^*\mid \mathcal{W}, \mathcal{P}_{<t}^*\right)
\end{equation}
where $\mathcal{P}_{<t}^*$ represents a partial demonstration path, $a_t^*$ is the expert action of $\mathcal{P}_{<t}^*$.

\paragraph{Object Grounding (OG)}
For tasks with object annotations, the object grounding~\cite{OG} loss is employed.
\begin{equation}
   L_{OG} = -\log p\left(o^* \mid \mathcal{W}, \mathcal{P}_T\right) 
\end{equation}
where $o^*$ refers to the object category at the agent's final destination $\mathcal{P}_T$ .

\subsubsection{Fine-tuning and Inference}
In the phase of fine-tuning, we employ a triad of loss functions: Single-step Action Prediction (SAP), Object Grounding (OG), and Turning Back Penalty (TBP), to meticulously guide and refine the learning process of the agent.
Distinct from the approach of utilizing a demonstration path in the pre-training phase, during fine-tuning, the model is supervised by a pseudo-interactive demonstrator.
This pseudo-interactive demonstrator operates by employing the shortest path algorithm at each decision point, selecting a node as the ensuing location. 
This selection is made such that it adheres to the criterion of minimizing the overall path length from the agent's current location to the target destination.
The cumulative loss function that governs the fine-tuning process is formulated as follows:
\begin{equation}
    L = \lambda_1 L_{SAP} + \lambda_2 L_{OG} + \lambda_3 L_{TBP}
\end{equation}
In this expression, $\lambda_1$, $\lambda_2$, and $\lambda_3$ serve as balance factors, ensuring a harmonious integration of the individual loss components.

For inference, the agent predicts one action at each step. 
If the action is not a stop action, it executes the action and moves to the appropriate position. Otherwise, the agent stops at the current position. 
When the number of actions executed by the agent exceeds the maximum number of actions specified, it is forced to stop and the node with the maximum stop probability is taken as the final position. 
When the agent stops, it will take the object with the maximum prediction score as the target.

\section{Experiments}\label{experiment}
\subsection{Datasets}
In this paper, we mainly focus on using relationships between objects to improve the ability of agent to perform visual-and-language navigation, so we choose two datasets with object annotations, REVERIE~\cite{REVERIE} and SOON~\cite{SOON}, to evaluate our model.
Additionally, we also present the results of our model on the R2R~\cite{R2R} dataset, which lacks object annotations.

\textbf{REVERIE} dataset mainly consists of instructions that describe target locations and objects of interest, averaging 21 words per instruction. 
It provides the agent with bounding boxes for each object in various panoramas. 
The agent must correctly identify and choose the right object bounding box at the end of its navigation. 
Paths demonstrated by experts in this dataset vary in length, ranging from 4 to 7 steps.

\textbf{SOON} dataset provides detailed instructions that accurately identify target rooms and objects, averaging 47 words in length. 
Unlike other datasets, SOON does not include predefined bounding boxes for objects. 
This requires the agent to predict objects' central locations within the panoramas.
To facilitate this, we utilize an automatic object detection approach as described in~\cite{BUTD}. 
This method helps us generate potential bounding boxes for the objects. 
The lengths of expert demonstrations in SOON are varied, ranging from 2 to 21 steps, with an average of approximately 9.5 steps.

\textbf{R2R} dataset encompasses a total of 21,567 words, with an average instruction length of 29 words. 
Due to the absence of object annotations in this dataset, we substitute object features with features from panoramic images.
In experiments, we solely train the agent using temporal object relations.

\subsection{Evaluation Metrics}
To assess the performance of our method in comparison to previous works, we adopt the conventional evaluation metrics for visual-and-language navigation task, as delineated in~\cite{R2R, REVERIE}. 
These metrics encompass: (1) Trajectory Length (TL)---the agent’s average path length in meters; (2) Navigation Error (NE)---average distance in meters between agent's final location and the target; (3) Success Rate (SR)---the percentage of instructions that are successfully executed, with an NE smaller than 3 meters; (4) Oracle SR (OSR)---SR given the oracle stop policy; (5) SPL---SR weighted by Path Length; (6) Remote Grounding Success (RGS)---the percentage of instructions that are executed successfully.; (7) RGSPL---RGS penalized by Path Length. Except for TL and NE, all metrics are higher the better.

\subsection{Implementation Details}
\subsubsection{Model Architectures}
For the extraction of object features within the REVERIE dataset, which conveniently offers bounding boxes, we employ the ViT-B/16 model~\cite{ViT}, pretrained on the ImageNet dataset. In contrast, when dealing with the SOON dataset, object bounding boxes are extracted using the BUTD object detector~\cite{BUTD} due to its absence of provided bounding boxes.
For the R2R dataset, we do not employ the spatial object relations module because of the unavailability of object annotations.

Further delving into our architecture, we incorporate a dual-scale graph transformer~\cite{DUET}. 
The specific configuration of this transformer includes setting the number of layers for the language encoder, panorama encoder, coarse-scale cross-modal encoder, and fine-scale cross-modal encoder to 9, 2, 4, and 4, respectively. 
The parameters for this segment of our model are initialized using the pretrained LXMERT model~\cite{LXMERT}. 

In the process of computing the relationship matrix $\mathcal{E}$, we assigned specific values to the parameters $k_1$, $k_2$, and $k_3$, setting them at 2, 2, and $5e^{-4}$, respectively. 
Furthermore, in order to compute the object relationship feature $\mathcal{Q}$, the weight parameters $\alpha_1$, $\alpha_2$, and $\alpha_3$ were meticulously initialized to 0.8, 0.1, and 0.1, respectively. 
Due to lacking spatial object relations module, we set $\alpha_1$, $\alpha_2$, and $\alpha_3$ as 0.8, 0.2, and 0 when using R2R dataset. 
Analogously, for the precise calculation of the loss function $L$, we established the values of the weight parameters $\lambda_1$, $\lambda_2$, and $\lambda_3$ at 1, 1, and 0.2, respectively.

\begin{table*}[!htbp]
 \centering
 \caption{Comparison with the state-of-the-art methods on the REVERIE dataset. The baseline indicates the replicated results of DUET.}
 \resizebox{\linewidth}{!}{\begin{tabular}{@{}c|cccc|cc|cccc|cc|cccc|cc@{}}
    \hline
    \multirow{3}{*}{Methods} & \multicolumn{6}{c|}{Val seen} &\multicolumn{6}{c|}{Val Unseen} & \multicolumn{6}{c}{Test Unseen}\\
    \cline{2-19}
    & \multicolumn{4}{c|}{Navigation} & \multicolumn{2}{c|}{Grounding} & \multicolumn{4}{c|}{Navigation} & \multicolumn{2}{c|}{Grounding} & \multicolumn{4}{c|}{Navigation} & \multicolumn{2}{c}{Grounding}\\
    \cline{2-19}
    & TL & OSR & SR & SPL & RGS & RGSPL & TL & OSR & SR & SPL & RGS & RGSPL & TL & OSR & SR & SPL & RGS & RGSPL\\
    \hline
    Seq2Seq~\cite{R2R} & 12.88 & 35.70 & 29.59 & 24.01 & 18.97 & 14.96 & 11.07 & 8.07 & 4.20 & 2.84 & 2.16 & 1.63 & 10.89 & 6.88 & 3.99 & 3.09 & 2.00 & 1.58\\
    RCM~\cite{RCM} & 10.70 & 29.44 & 23.33 & 21.82 & 13.23 & 15.36 & 11.98 & 14.23 & 9.39 & 6.97 & 4.89 & 3.89 & 10.60 & 11.68 & 7.84 & 6.67 & 3.67 & 3.14\\
    VLNBERT~\cite{VLNBERT} & 13.44 & 53.90 & 51.79 & 47.96 & 38.23 & 35.61 & 16.78 & 35.02 & 30.67 & 24.90 & 18.77 & 15.27 & 15.68 & 32.91 & 29.61 & 23.99 &  16.50 & 13.51\\
    AirBERT~\cite{AirBERT} & 15.16 & 49.98 & 47.01 & 42.34 & 32.75 & 30.01 & 18.71 & 34.51 & 27.89 & 21.88 & 18.23 & 14.18 & 17.91 & 34.20 & 30.28 & 23.61 &  16.83 & 13.28\\
    HOP~\cite{HOP} & 13.80 & 54.88 & 53.76 & 47.19 & 38.65 & 33.85 & 16.46 & 36.24 & 31.78 & 26.11 & 18.85 & 15.73 & 16.38 & 33.06 & 30.17 & 24.34 & 17.69 & 14.34\\
    HAMT~\cite{SAP} & 12.79 & 47.65 & 43.29 & 40.19 & 27.20 & 15.18 & 14.08 & 36.84 & 32.95 & 30.20 & 18.92 & 17.28 & 13.62 & 33.41 & 30.40 & 26.67 & 14.88 & 13.08\\
    CKR~\cite{CKR} & 12.16 & 61.91 & 57.27 & 53.57 & 39.07 & - & 26.26 & 31.44 & 19.14 & 11.84 & 11.45 & - & 22.46 & 30.40 & 22.00 & 14.25 & 11.60 & -\\
    DUET~\cite{DUET} & 13.86 & 73.68 & 71.75 & 63.94 & 57.41 & 51.14 & 22.11 & 51.07 & 46.98 & 33.73 & 32.15 & 23.03 & 21.30 & 56.91 & 52.51 & 36.06 & 31.88 & 22.06\\
    KERM~\cite{KERM} & 12.84 & 79.20 & 76.88 & 70.45 & 61.00 & 56.07 & 21.85 & 55.21 & 50.44 & 35.38 & 34.51 & 24.45 & 17.32 & 57.58 & 52.43 & 39.21 & 32.39 & 23.64\\
    BEVBert~\cite{BEVBert} & - & - & - & - & - & - & - & 56.40 & \textbf{51.78} & 36.37 & 34.71 & 24.44 & - & 57.26 & 52.81 & 36.41 & 32.06 & 22.09\\
    GridMM~\cite{GridMM} & - & - & - & - & - & - & 23.20 & \textbf{57.48} & 51.37 & 36.47 & 34.57 & 24.56 & 19.97 & 59.55 & 53.13 & 36.60 & 34.87 & 23.45\\
    \hline
    Baseline & 13.84 & 73.79 & 71.68 & 63.90 & 57.34 & 51.11 & 22.12 & 51.09 & 46.98 & 33.75 & 32.15 & 23.05 & 18.19 & 55.31 & 50.67 & 36.27 & 31.87 & 22.65\\
    Ours & 14.00 & \textbf{83.06} & \textbf{80.46} & \textbf{73.12} & \textbf{64.02} & \textbf{58.29} & 22.00 & 55.55 & 50.30 & \textbf{36.84} & \textbf{35.27} & \textbf{25.98} & 20.25 & \textbf{61.08} & \textbf{55.31} & \textbf{40.37} & \textbf{35.16} & \textbf{24.99}\\
    \hline
 \end{tabular}}
 \label{table1}
\end{table*}

\begin{table*}[!htbp]
 \centering
 \caption{Comparison with the state-of-the-art methods on the SOON dataset. The baseline indicates the replicated results of DUET.}
 \resizebox{0.8\linewidth}{!}{\begin{tabular}{@{}c|ccccc|ccccc@{}}
    \hline
    \multirow{2}{*}{Methods} & \multicolumn{5}{c|}{Val Unseen} & \multicolumn{5}{c}{Test Unseen}\\
    \cline{2-11}
    & TL & OSR & SR & SPL & RGSPL & TL & OSR & SR & SPL & RGSPL\\
    \hline
    GBE~\cite{GBE} & 28.96 & 28.54 & 19.52 & 13.34 & 1.16 & 27.88 & 21.45 & 12.90 & 9.23 & 0.45 \\
    DUET~\cite{DUET} & 36.20 & 50.91 & 36.28 & 22.58 & 3.75 & 41.83 & 43.00 & 33.44 & 21.42  & 4.17\\
    KERM~\cite{KERM} & 35.83 & 51.62 & 38.05 & 23.16 & 4.04 & - & - & - & - & - \\
    GridMM~\cite{GridMM} & 38.92 & 53.39 & 37.46 & 24.81 & 3.91 & 46.20 & \textbf{48.02} & 36.27 & 21.25 & 4.15 \\
    \hline
    Baseline & 36.18 & 50.88 & 36.17 & 22.54 & 3.75 & 40.73 & 41.74 & 32.05 & 20.79  & 4.55\\
    Ours & 38.75 & \textbf{55.46} & \textbf{40.12} & \textbf{26.00} & \textbf{5.04} & 40.05 & 47.73 & \textbf{37.09} & \textbf{23.60} & \textbf{6.35}\\
    \hline
 \end{tabular}}
 \label{table2}
\end{table*}

\subsubsection{Training Details}
For pretraining, we set the batch size as 32 using 1 NVIDIA RTX3090 GPU. 
For the REVERIE dataset, we combine the original dataset with augmented data synthesized by DUET~\cite{DUET} to pretrain our model with 100k iterations. 
Then we fine-tune the pretrained model with the batch size of 16 for 20k iterations on 1 NVIDIA RTX3090 GPU. 
For the SOON dataset, we only use the original data with automatically cleaned object bounding boxes, sharing the same settings in DUET~\cite{DUET}. 
We pretrain the model with 40k iterations. 
Then we fine-tune the pretrained model with the batch size of 4 for 40k iterations on 1 NVIDIA RTX3090 GPU. 
For the R2R dataset, additional augmented R2R data in~\cite{pretrain3} is used in pretraining.
We pretrain the model for 200k iterations with batch size of 64 and then fine-tune it for 20k iterations with batch size of 8.

\subsection{Performance Comparison}
The results of our method on the REVERIE and SOON datasets are depicted in \cref{table1} and \cref{table2}, respectively.
Across most metrics on these two datasets, our approach achieves superior performance. 
Specifically, in the test split of REVERIE, as detailed in \cref{table1}, our method achieves notable improvements over the baseline: 4.64\% on SR, 4.10\% on SPL, 3.29\% on RGS, and 2.34\% on RGSPL. 
This substantial enhancement underscores the robustness and efficacy of our technique. 
Moreover, even when compared to the current state-of-the-art method, GridMM\cite{GridMM}, our method still shows advancements of 2.18\%, 3.77\%, 0.29\%, and 1.54\% on SR, SPL, RGS, and RGSPL, respectively, highlighting the superior capability of our approach. 
As indicated in~\cref{table2}, on the more intricate SOON dataset, our method also manifests exceptional performance, surpassing the current state-of-the-art. 
This underscores our method's proficiency in grasping inter-object relations, thereby enhancing the agent's navigational prowess.

In our study, we also conducted experiments on the R2R dataset, which lacks object annotations. 
As shown in~\cref{table3}, the absence of explicit object information impeded the agent's ability to accurately learn the relationships between objects in the navigation environment.
Our method does not exhibit significant improvements over other methods in both the val unseen and test splits across various metrics. 
Interestingly, we observes a remarkable phenomenon: on the val seen split, our method significantly outperforms both the baseline and other approaches. 
This can be attributed to the agent's repeated exposure to the panoramas in the val seen split during training. 
Despite the lack of explicit object annotations, the agent often manages to infer the presence and relationships of objects based on these panoramic features.
These findings further illustrate the vital role of object relationships in enabling an agent to accurately complete navigation tasks. 
This aspect of our research highlights the significance of understanding and integrating object interactions within the navigational context for improved agent performance.

\subsection{Ablation Study}
\subsubsection{Ablation of object relations.}
To explore the effects of our proposed modules, TOR and SOR, on the agent's navigation skills, we integrated them separately into the baseline method. 
We then conducted experiments with these integrations on the val unseen split of the REVERIE and SOON datasets. 
As illustrated in \cref{table4}, both TOR and SOR notably enhance the navigation performance of the agent. 
However, we observed that SOR contributes to a more modest improvement in navigation performance compared to TOR. 
This is attributed to the limited scale of the current datasets used for training the agent.
Solely relying on the objects observed at each position within the environment is inadequate to accurately capture the relationships among different objects. 
When both modules are employed in tandem, the navigational prowess of the agent is further amplified.

\begin{table*}[!htbp]
 \centering
 \caption{Comparison with the state-of-the-art methods on the R2R dataset. The baseline indicates the replicated results of DUET.}
 \resizebox{0.9\linewidth}{!}{\begin{tabular}{@{}c|cccc|cccc|cccc@{}}
    \hline
    \multirow{2}{*}{Methods} & \multicolumn{4}{c|}{Val seen} & \multicolumn{4}{c|}{Val Unseen} & \multicolumn{4}{c}{Test Unseen}\\
    \cline{2-13}
    & TL & NE & SR & SPL & TL & NE & SR & SPL & TL & NE & SR & SPL\\
    \hline
    Seq2Seq~\cite{R2R} & 11.33 & 6.01 & 39 & - & 8.39 & 7.81 & 22 & - & 8.13 & 7.85 & 20 & 18\\
    RCM~\cite{RCM} & - & 3.33 & 70 & 67 & - & 5.28 & 55 & 50 & - & 5.15 & 55 & 51\\
    PREVALENT~\cite{pretrain3} & 10.32 & 3.67 & 69 & 65 & 10.19 & 4.71 & 58 & 53 & 10.51 & 5.30 & 54 & 51\\
    EntityGraph~\cite{LVER} & 10.13 & 3.47 & 67 & 65 & 9.99 & 4.73 & 57 & 53 & 10.29 & 4.75 & 55 & 52\\
    VLNBERT~\cite{VLNBERT} & 11.13 & 2.90 & 72 & 68 & 12.01 & 3.93 & 63 & 57 & 12.35 & 4.09 & 63 & 57\\
    AirBERT~\cite{AirBERT} & 11.09 & 2.68 & 75 & 70 & 11.78 & 4.01 & 62 & 56 & 12.41 & 4.13 & 62 & 67\\
    HOP~\cite{HOP} & 11.26 & 2.72 & 75 & 70 & 12.27 & 3.80 & 64 & 57 & 12.68 & 3.83 & 64 & 59\\
    HAMT~\cite{SAP} & - & - & - & - & 11.46 & 2.29 & 66 & 61 & 12.27 & 3.93 & 65 & 60\\
    DUET~\cite{DUET} & - & - & - & - & 13.94 & 3.31 & 72 & 60 & 14.73 & 3.65 & 69 & 59\\
    KERM~\cite{KERM} & 12.16 & 2.19 & 79.73 & 73.79 & 13.54 & 3.22 & 71.95 & 60.91 & 14.60 & 3.61 & 69.73 & 59.25\\
    BEVBert~\cite{BEVBert} & - & - & - & - & - & \textbf{2.81} & 75 & 64 & - & \textbf{3.13} & 73 & 62\\
    GridMM~\cite{GridMM} & - & - & - & - & 13.27 & 2.83 & \textbf{75} & \textbf{64} & 14.43 & 3.35 & \textbf{73} & 62\\
    \hline
    Baseline & 13.41 & 2.35 & 79 & 72 & 13.92 & 3.25 & 71 & 60 & 15.2 & 3.42 & 70 & 60\\
    Ours & 12.67 & \textbf{2.05} & \textbf{82} & \textbf{76} & 13.57 & 3.05 & 71 & 61 & 15.4 & 3.27 & 72 & \textbf{62}\\
    \hline
 \end{tabular}}
 \label{table3}
\end{table*}

\begin{table}[tbp]
 \centering
 \caption{Ablation of the object relations on the val unseen split of REVERIE and SOON. In this table, TOR and EOR respectively represent the trajectory-object relationships module and the environment-object relationships module. All experiments are conducted without utilizing TBP loss function. }
 \resizebox{\linewidth}{!}{\begin{tabular}{@{}ccc|cccccc@{}}
    \hline
     Dataset & TOR & EOR & TL & OSR & SR & SPL & RGS & RGSPL \\
     \hline
     \multirow{4}{*}{REVERIE} & $\times$ & $\times$ & 22.12 & 51.09 & 46.98 & 33.75 & 32.15 & 23.05 \\
     & $\times$ & \checkmark & 22.31 & 52.85 & 48.08 & 33.77 & 32.55 & 23.07 \\
     & \checkmark & $\times$ & 23.07 & \textbf{55.44} & 49.22 & 33.32 & 33.46 & 22.90 \\
     & \checkmark & \checkmark & 23.52 & 54.47 & \textbf{49.64} & \textbf{34.56} & \textbf{34.05} & \textbf{23.41} \\
     \hline
     \multirow{4}{*}{SOON} & $\times$ & $\times$ & 36.18 & 50.88 & 36.17 & 22.54 & 6.02 & 3.75 \\
     & $\times$ & \checkmark & 37.07 & 51.92 & 38.05 & \textbf{25.31} & 5.46 & 3.51 \\
     & \checkmark & $\times$ & 36.46 & 54.28 & 39.09 & 25.13 & 7.08 & \textbf{4.61} \\
     & \checkmark & \checkmark & 39.91 & \textbf{56.49} & \textbf{40.71} & 24.98 & \textbf{7.37} & 4.56 \\
     \hline
 \end{tabular}}
 \label{table4}
\end{table}

\begin{figure}[!htbp]
	\centering
	\includegraphics[width=\linewidth]{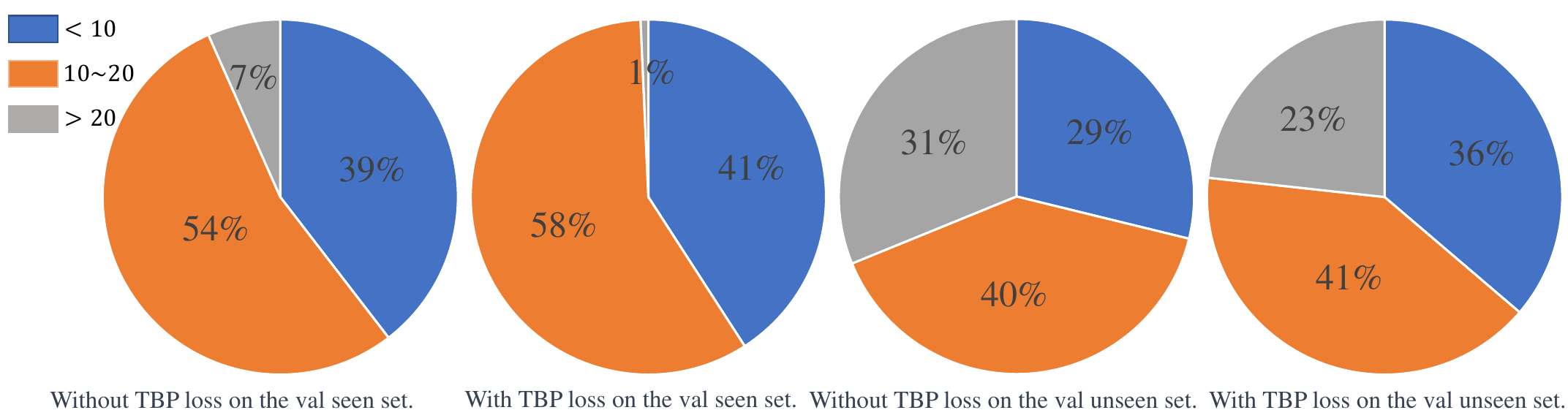}
	
	\caption{The distribution of trajectory lengths predicted on the val seen and val unseen splits of REVERIE dataset.}
 
	\label{pathLength}
\end{figure}

\begin{table}[!htbp]
 \centering
 \caption{Ablation of TBP loss on the val unseen split of REVERIE and SOON. Both the TOR and EOR modules are employed in these experiments. }
 \resizebox{\linewidth}{!}{\begin{tabular}{@{}cc|cccccc@{}}
    \hline
     Dataset & TBP & TL & OSR & SR & SPL & RGS & RGSPL \\
     \hline
     \multirow{2}{*}{REVERIE} & $\times$ & 23.52 & 54.47 & 49.64 & 34.56 & 34.05 & 23.41 \\
     & \checkmark & 22.00 & \textbf{55.55} & \textbf{50.30} & \textbf{36.79} & \textbf{35.27} & \textbf{25.98} \\
     \hline
     \multirow{2}{*}{SOON} &$\times$ & 39.91 & \textbf{56.49} & \textbf{40.71} & 24.98 & 7.37 & 4.56 \\
     & \checkmark & 38.75 & 55.46 & 40.12 & \textbf{26.00} & \textbf{7.67} & \textbf{5.04} \\
     \hline
 \end{tabular}}
 \label{table5}
\end{table}

\subsubsection{Ablation of TBP loss.}
In~\cref{table4}, we observes an intriguing phenomenon. 
The inclusion of object relationships does indeed significantly enhance the agent's success rates in navigation (OSR, SR, and RGS). 
However, this enhancement also leads to an increase in the trajectory length (TL) of the agent. 
As a result, there is no pronounced improvement in metrics such as SPL and RGSPL.
This suggests that the object relationship module leads the agent to engage in excessive redundant exploration, resulting in elongated navigation paths. 
Upon integrating the TBP loss function, we observes a significant reduction in the agent's revisits to the same location. 
This is illustrated in \cref{table5}. 
This change leads to a more efficient task completion and shows clear improvements in metrics such as TL, SPL.

Additionally, \cref{pathLength} depicts the distribution of navigational path lengths corresponding to successful navigation instances (specifically when SR is 1) for both the val seen and val unseen splits of the REVERIE dataset, comparing scenarios with and without the integration of the TBP loss function.
The figure reveals that with the TBP loss, there's a notable increase in the proportion of agent paths falling between 0 and 10, while paths longer than 20 significantly diminish. 
This solidly validates that our TBP loss function effectively curtails the navigation path length of the agent.

\subsubsection{Punish turning back during inference.}
\begin{table}[!tbp]
 \centering
 \caption{The results of punishing turning back during inference. $\xi = i$ means dividing the probability of the agent reaching a certain position by $i$.}
 \resizebox{\linewidth}{!}{\begin{tabular}{@{}c|c|cccccc@{}}
    \hline
     \text{No.} & $\xi$ & TL & OSR & SR & SPL & RGS & RGSPL \\
     \hline
     1 & 0.5 & 28.10 & \textbf{59.24} & 38.00 & 22.17 & 27.78 & 16.66 \\
     2 & 1 & 22.00 & 55.55 & \textbf{50.30} & \textbf{36.79} & \textbf{35.27} &
     \textbf{25.98} \\
     3 & 2 & 22.53 &  51.32 & 47.71 & 32.91 & 32.06 & 22.22\\
     4 & 4 & 22.01 & 49.53 & 46.49 & 32.35 & 31.47 & 21.91\\
     5 & 6 & 21.90 & 48.79 & 45.81 & 32.03 & 31.10 & 21.75\\
     6 & 8 & 21.84 & 48.59 & 45.61 & 31.93 & 30.99 & 21.73\\
     \hline
 \end{tabular}}
 \label{table6}
\end{table}

It is also a intuitive way punish turning back during inference.
To assess whether penalizing the agent's repetitive visiting behavior during inference improves its navigational abilities, we have designed several experiments.
The experimental outcomes are presented in~\cref{table6}.

Our findings indicate that penalizing the agent's repetitive visiting behavior during inference does not enhance its navigational performance. 
As evidenced in the last five rows of~\cref{table6}, the navigation success rate of the agent decreases with increasing penalty intensity. 
This decline in performance is attributed to the fact that such penalties during inference prevent the agent from correcting its navigational errors. 
Additionally, the first two rows of~\cref{table6} reveal that when repetitive visits by the agent are encouraged, there is a more significant drop in navigational ability. 
This is due to the agent engaging in more unproductive exploration, substantially increasing the length of the navigational path.

\subsection{Qualitative Results}

\begin{figure*}[!htbp]
    \centering
    \begin{subfigure}[t]{0.6\textwidth}
        \centering
        \includegraphics[width=\linewidth]{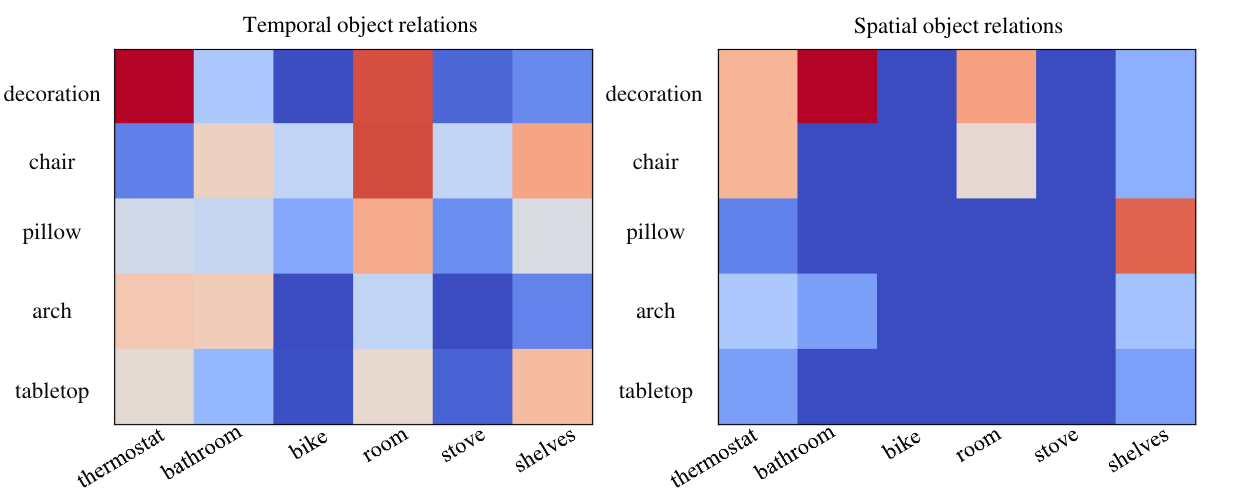} 
        \caption{The figures show the attention maps of the temporal object relations and spatial object relations. The attention is indicated from low to high with the color blue to red.}
        \label{heatmap}
    \end{subfigure}%
    \hfill
    \begin{subfigure}[t]{0.34\textwidth}
        \centering
        \includegraphics[width=\linewidth]{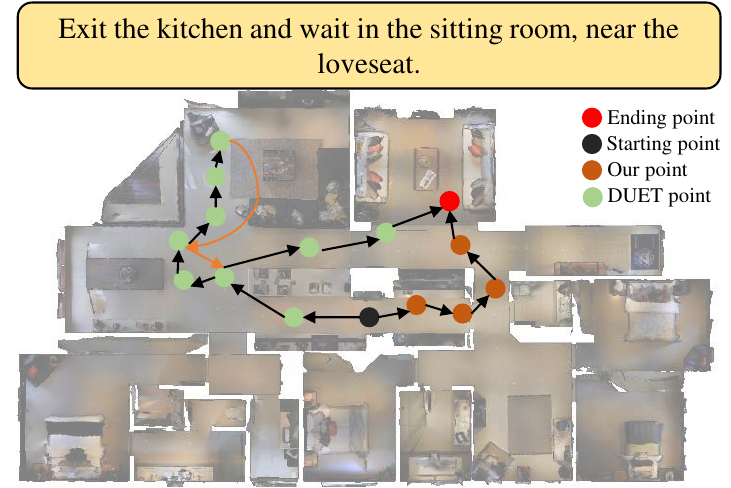} 
        \caption{In this figure, orange lines emphasize instances where the agent backtracks on its route.}
        \label{path}
    \end{subfigure}
    
    \caption{Visualization of attention maps and navigation examples}
    \label{qualitative}
\end{figure*}

\subsubsection{Visualization of object relations.}
To demonstrate that our method can accurately learn the various relationships between objects, we visualized attention heatmaps of different object interactions after training. 
\cref{heatmap} showcases the object relationships learned by both the TOR and SOR modules. 
From this, we observes that both modules are capable of accurately modeling object relationships. 
The spatial object relations provide a detailed and comprehensive depiction of inherent connections between objects, such as chairs and rooms, or decor and thermostats being closely associated. 
The spatial object relations supplement the temporal object relations.
Certain information not captured by the TOR module is correctly learned in SOR. 
For instance, the environmental object relationships suggest a connection between decor and bathroom, which aligns with real-world scenarios but is not precisely represented in the trajectory-object relationships.

\subsubsection{Visualization of the navigation trajectories.}
To elucidate the effectiveness of our proposed approach, we have rendered a comparative visualization of the navigation trajectories generated by both DUET and our method. 
As illustrated in~\cref{path}, both techniques can accurately reach navigation targets.
However, DUET tends to involve the agent in excessive exploratory actions. 
This often results in repeated visits to the same places, hindering navigational efficiency.
Conversely, our method substantially diminishes the agent's inclination to backtrack, facilitating the selection of more direct routes that expedite the completion of the navigation tasks.
Furthermore, our analysis reveals that at the onset of navigation, our method opted for a proximal route, in contrast to DUET which embarked on a relatively longer path with a greater number of actions. 
This indicates that our method can effectively understand the relationships between objects encountered by the agent and the specified targets. 
Consequently, it identifies more efficient paths. 
This aids in completing navigational tasks more effectively.

\section{Conclusion}\label{conclusion}
In this study, we introduced Temporal-Spatial Object Relations Modules and a Turning Back Penalty  (TBP) loss function that together enhance agent navigation. 
By learning the connections between various objects, the agent can more effectively complete navigation tasks. 
The application of the TBP loss function successfully prevents repetitive visits to the same location by the agent, thereby enhancing navigational efficiency.
It is noteworthy that the object relationships we model might not be entirely accurate because of the limited datasets.
Moving forward, our future endeavors will focus on devising more sophisticated object relationship modeling techniques, expanding dataset scales, and honing the precision and efficiency of navigational tasks.

\bibliographystyle{IEEEtran}
\bibliography{main}

\begin{thebibliography}{10}
\providecommand{\url}[1]{#1}
\csname url@samestyle\endcsname
\providecommand{\newblock}{\relax}
\providecommand{\bibinfo}[2]{#2}
\providecommand{\BIBentrySTDinterwordspacing}{\spaceskip=0pt\relax}
\providecommand{\BIBentryALTinterwordstretchfactor}{4}
\providecommand{\BIBentryALTinterwordspacing}{\spaceskip=\fontdimen2\font plus
\BIBentryALTinterwordstretchfactor\fontdimen3\font minus
  \fontdimen4\font\relax}
\providecommand{\BIBforeignlanguage}[2]{{%
\expandafter\ifx\csname l@#1\endcsname\relax
\typeout{** WARNING: IEEEtran.bst: No hyphenation pattern has been}%
\typeout{** loaded for the language `#1'. Using the pattern for}%
\typeout{** the default language instead.}%
\else
\language=\csname l@#1\endcsname
\fi
#2}}
\providecommand{\BIBdecl}{\relax}
\BIBdecl

\bibitem{SFT}
M.~Nawaz, J.~K.-T. Tang, K.~Bibi, S.~Xiao, H.-P. Ho, and W.~Yuan, ``Robust
  cognitive capability in autonomous driving using sensor fusion techniques: A
  survey,'' \emph{IEEE Transactions on Intelligent Transportation Systems},
  vol.~25, no.~5, pp. 3228--3243, 2024.

\bibitem{ADP}
C.~Sun, R.~Zhang, Y.~Lu, Y.~Cui, Z.~Deng, D.~Cao, and A.~Khajepour, ``Toward
  ensuring safety for autonomous driving perception: Standardization progress,
  research advances, and perspectives,'' \emph{IEEE Transactions on Intelligent
  Transportation Systems}, vol.~25, no.~5, pp. 3286--3304, 2024.

\bibitem{AirBERT}
P.-L. Guhur, M.~Tapaswi, S.~Chen, I.~Laptev, and C.~Schmid, ``Airbert:
  In-domain pretraining for vision-and-language navigation,'' in
  \emph{Proceedings of the IEEE/CVF International Conference on Computer Vision
  (ICCV)}, October 2021, pp. 1634--1643.

\bibitem{R2R}
P.~Anderson, Q.~Wu, D.~Teney, J.~Bruce, M.~Johnson, N.~Sünderhauf, I.~Reid,
  S.~Gould, and A.~van~den Hengel, ``Vision-and-language navigation:
  Interpreting visually-grounded navigation instructions in real
  environments,'' in \emph{2018 IEEE/CVF Conference on Computer Vision and
  Pattern Recognition}, 2018, pp. 3674--3683.

\bibitem{DIAG}
W.~Zhu, Y.~Qi, P.~Narayana, K.~Sone, S.~Basu, X.~Wang, Q.~Wu, M.~Eckstein, and
  W.~Y. Wang, ``Diagnosing vision-and-language navigation: What really
  matters,'' in \emph{Proceedings of the 2022 Conference of the North American
  Chapter of the Association for Computational Linguistics: Human Language
  Technologies}, 2022, pp. 5981--5993.

\bibitem{REVERIE}
Y.~Qi, Q.~Wu, P.~Anderson, X.~Wang, W.~Y. Wang, C.~Shen, and A.~van~den Hengel,
  ``Reverie: Remote embodied visual referring expression in real indoor
  environments,'' in \emph{2020 IEEE/CVF Conference on Computer Vision and
  Pattern Recognition (CVPR)}, 2020, pp. 9979--9988.

\bibitem{IC}
W.~Kang, J.~Mun, S.~Lee, and B.~Roh, ``Noise-aware learning from web-crawled
  image-text data for image captioning,'' in \emph{Proceedings of the IEEE/CVF
  International Conference on Computer Vision}, 2023, pp. 2942--2952.

\bibitem{VQA}
Y.~Zhang, C.-H. Ho, and N.~Vasconcelos, ``Toward unsupervised realistic visual
  question answering,'' in \emph{Proceedings of the IEEE/CVF International
  Conference on Computer Vision (ICCV)}, October 2023, pp. 15\,613--15\,624.

\bibitem{VisDial}
J.~Qi, Y.~Niu, J.~Huang, and H.~Zhang, ``Two causal principles for improving
  visual dialog,'' in \emph{2020 IEEE/CVF Conference on Computer Vision and
  Pattern Recognition (CVPR)}, 2020, pp. 10\,857--10\,866.

\bibitem{RE}
Y.~Qiao, C.~Deng, and Q.~Wu, ``Referring expression comprehension: A survey of
  methods and datasets,'' \emph{IEEE Transactions on Multimedia}, vol.~23, pp.
  4426--4440, 2021.

\bibitem{VLNBERT}
Y.~Hong, Q.~Wu, Y.~Qi, C.~Rodriguez-Opazo, and S.~Gould, ``Vln bert: A
  recurrent vision-and-language bert for navigation,'' in \emph{2021 IEEE/CVF
  Conference on Computer Vision and Pattern Recognition (CVPR)}, 2021, pp.
  1643--1653.

\bibitem{speaker}
D.~Fried, R.~Hu, V.~Cirik, A.~Rohrbach, J.~Andreas, L.-P. Morency,
  T.~Berg-Kirkpatrick, K.~Saenko, D.~Klein, and T.~Darrell, ``Speaker-follower
  models for vision-and-language navigation,'' \emph{Advances in Neural
  Information Processing Systems}, vol.~31, 2018.

\bibitem{SOTA}
A.~Moudgil, A.~Majumdar, H.~Agrawal, S.~Lee, and D.~Batra, ``Soat: A scene- and
  object-aware transformer for vision-and-language navigation,'' in
  \emph{Advances in Neural Information Processing Systems}, M.~Ranzato,
  A.~Beygelzimer, Y.~Dauphin, P.~Liang, and J.~W. Vaughan, Eds., vol.~34.\hskip
  1em plus 0.5em minus 0.4em\relax Curran Associates, Inc., 2021, pp.
  7357--7367.

\bibitem{BTED}
H.~Tan, L.~Yu, and M.~Bansal, ``Learning to navigate unseen environments: Back
  translation with environmental dropout,'' in \emph{Proceedings of the 2019
  Conference of the North American Chapter of the Association for Computational
  Linguistics: Human Language Technologies, Volume 1 (Long and Short Papers)},
  2019, pp. 2610--2621.

\bibitem{ADAPT}
B.~Lin, Y.~Zhu, Z.~Chen, X.~Liang, J.~Liu, and X.~Liang, ``Adapt:
  Vision-language navigation with modality-aligned action prompts,'' in
  \emph{2022 IEEE/CVF Conference on Computer Vision and Pattern Recognition
  (CVPR)}, 2022, pp. 15\,375--15\,385.

\bibitem{DUET}
S.~Chen, P.-L. Guhur, M.~Tapaswi, C.~Schmid, and I.~Laptev, ``Think global, act
  local: Dual-scale graph transformer for vision-and-language navigation,'' in
  \emph{2022 IEEE/CVF Conference on Computer Vision and Pattern Recognition
  (CVPR)}, 2022, pp. 16\,516--16\,526.

\bibitem{KERM}
X.~Li, Z.~Wang, J.~Yang, Y.~Wang, and S.~Jiang, ``Kerm: Knowledge enhanced
  reasoning for vision-and-language navigation,'' in \emph{2023 IEEE/CVF
  Conference on Computer Vision and Pattern Recognition (CVPR)}, 2023, pp.
  2583--2592.

\bibitem{GridMM}
Z.~Wang, X.~Li, J.~Yang, Y.~Liu, and S.~Jiang, ``Gridmm: Grid memory map for
  vision-and-language navigation,'' in \emph{Proceedings of the IEEE/CVF
  International Conference on Computer Vision (ICCV)}, October 2023, pp.
  15\,625--15\,636.

\bibitem{ORG}
H.~Du, X.~Yu, and L.~Zheng, ``Learning object relation graph and tentative
  policy for visual navigation,'' in \emph{Computer Vision -- ECCV 2020},
  A.~Vedaldi, H.~Bischof, T.~Brox, and J.-M. Frahm, Eds.\hskip 1em plus 0.5em
  minus 0.4em\relax Cham: Springer International Publishing, 2020, pp. 19--34.

\bibitem{DOA}
R.~Dang, Z.~Shi, L.~Wang, Z.~He, C.~Liu, and Q.~Chen, ``Unbiased directed
  object attention graph for object navigation,'' in \emph{Proceedings of the
  30th ACM International Conference on Multimedia}, 2022, pp. 3617--3627.

\bibitem{DAT}
R.~Dang, L.~Wang, Z.~He, S.~Su, J.~Tang, C.~Liu, and Q.~Chen, ``Search for or
  navigate to? dual adaptive thinking for object navigation,'' in
  \emph{Proceedings of the IEEE/CVF International Conference on Computer Vision
  (ICCV)}, October 2023, pp. 8250--8259.

\bibitem{SEvol}
J.~Chen, C.~Gao, E.~Meng, Q.~Zhang, and S.~Liu, ``Reinforced structured
  state-evolution for vision-language navigation,'' in \emph{Proceedings of the
  IEEE/CVF Conference on Computer Vision and Pattern Recognition (CVPR)}, June
  2022, pp. 15\,450--15\,459.

\bibitem{BPOD}
C.-W. Kuo and Z.~Kira, ``Beyond a pre-trained object detector: Cross-modal
  textual and visual context for image captioning,'' in \emph{Proceedings of
  the IEEE/CVF Conference on Computer Vision and Pattern Recognition}, 2022,
  pp. 17\,969--17\,979.

\bibitem{VG}
R.~Krishna, Y.~Zhu, O.~Groth, J.~Johnson, K.~Hata, J.~Kravitz, S.~Chen,
  Y.~Kalantidis, L.-J. Li, D.~A. Shamma \emph{et~al.}, ``Visual genome:
  Connecting language and vision using crowdsourced dense image annotations,''
  \emph{International journal of computer vision}, vol. 123, pp. 32--73, 2017.

\bibitem{Conceptnet}
R.~Speer, J.~Chin, and C.~Havasi, ``Conceptnet 5.5: An open multilingual graph
  of general knowledge,'' in \emph{Proceedings of the AAAI conference on
  artificial intelligence}, vol.~31, 2017.

\bibitem{CKR}
C.~Gao, J.~Chen, S.~Liu, L.~Wang, Q.~Zhang, and Q.~Wu, ``Room-and-object aware
  knowledge reasoning for remote embodied referring expression,'' in
  \emph{Proceedings of the IEEE/CVF Conference on Computer Vision and Pattern
  Recognition}, 2021, pp. 3064--3073.

\bibitem{VELMA}
R.~{Schumann}, W.~{Zhu}, W.~{Feng}, T.-J. {Fu}, S.~{Riezler}, and W.~Y. {Wang},
  ``{VELMA: Verbalization Embodiment of LLM Agents for Vision and Language
  Navigation in Street View},'' \emph{arXiv e-prints}, p. arXiv:2307.06082,
  Jul. 2023.

\bibitem{NavGPT}
G.~{Zhou}, Y.~{Hong}, and Q.~{Wu}, ``{NavGPT: Explicit Reasoning in
  Vision-and-Language Navigation with Large Language Models},'' \emph{arXiv
  e-prints}, p. arXiv:2305.16986, May 2023.

\bibitem{LM-Nav}
D.~Shah, B.~Osi\'nski, b.~ichter, and S.~Levine, ``Lm-nav: Robotic navigation
  with large pre-trained models of language, vision, and action,'' in
  \emph{Proceedings of The 6th Conference on Robot Learning}, ser. Proceedings
  of Machine Learning Research, K.~Liu, D.~Kulic, and J.~Ichnowski, Eds., vol.
  205.\hskip 1em plus 0.5em minus 0.4em\relax PMLR, 14--18 Dec 2023, pp.
  492--504.

\bibitem{perception}
H.~Wang, W.~Wang, W.~Liang, S.~C. Hoi, J.~Shen, and L.~V. Gool, ``Active
  perception for visual-language navigation,'' \emph{International Journal of
  Computer Vision}, vol. 131, no.~3, pp. 607--625, 2023.

\bibitem{SOON}
F.~Zhu, X.~Liang, Y.~Zhu, Q.~Yu, X.~Chang, and X.~Liang, ``Soon: Scenario
  oriented object navigation with graph-based exploration,'' in \emph{2021
  IEEE/CVF Conference on Computer Vision and Pattern Recognition (CVPR)}, 2021,
  pp. 12\,684--12\,694.

\bibitem{Reinforced}
X.~Wang, Q.~Huang, A.~Celikyilmaz, J.~Gao, D.~Shen, Y.-F. Wang, W.~Y. Wang, and
  L.~Zhang, ``Reinforced cross-modal matching and self-supervised imitation
  learning for vision-language navigation,'' in \emph{Proceedings of the
  IEEE/CVF conference on computer vision and pattern recognition}, 2019, pp.
  6629--6638.

\bibitem{LVER}
Y.~Hong, C.~Rodriguez, Y.~Qi, Q.~Wu, and S.~Gould, ``Language and visual entity
  relationship graph for agent navigation,'' \emph{Advances in Neural
  Information Processing Systems}, vol.~33, pp. 7685--7696, 2020.

\bibitem{SSM}
H.~Wang, W.~Wang, W.~Liang, C.~Xiong, and J.~Shen, ``Structured scene memory
  for vision-language navigation,'' in \emph{Proceedings of the IEEE/CVF
  conference on Computer Vision and Pattern Recognition}, 2021, pp. 8455--8464.

\bibitem{HOP}
Y.~Qiao, Y.~Qi, Y.~Hong, Z.~Yu, P.~Wang, and Q.~Wu, ``Hop: History-and-order
  aware pretraining for vision-and-language navigation,'' in \emph{2022
  IEEE/CVF Conference on Computer Vision and Pattern Recognition (CVPR)}, 2022,
  pp. 15\,397--15\,406.

\bibitem{LU}
S.~Chen, P.-L. Guhur, M.~Tapaswi, C.~Schmid, and I.~Laptev, ``Learning from
  unlabeled 3d environments for vision-and-language navigation,'' in
  \emph{European Conference on Computer Vision}.\hskip 1em plus 0.5em minus
  0.4em\relax Springer, 2022, pp. 638--655.

\bibitem{RCM}
F.~Zhu, Y.~Zhu, X.~Chang, and X.~Liang, ``Vision-language navigation with
  self-supervised auxiliary reasoning tasks,'' in \emph{2020 IEEE/CVF
  Conference on Computer Vision and Pattern Recognition (CVPR)}, 2020, pp.
  10\,009--10\,019.

\bibitem{Regretful}
C.-Y. Ma, Z.~Wu, G.~AlRegib, C.~Xiong, and Z.~Kira, ``The regretful agent:
  Heuristic-aided navigation through progress estimation,'' in
  \emph{Proceedings of the IEEE/CVF conference on Computer Vision and Pattern
  Recognition}, 2019, pp. 6732--6740.

\bibitem{pretrain3}
W.~Hao, C.~Li, X.~Li, L.~Carin, and J.~Gao, ``Towards learning a generic agent
  for vision-and-language navigation via pre-training,'' \emph{2020 IEEE/CVF
  Conference on Computer Vision and Pattern Recognition (CVPR)}, pp.
  13\,134--13\,143, 2020.

\bibitem{PRESS}
X.~Li, C.~Li, Q.~Xia, Y.~Bisk, A.~Celikyilmaz, J.~Gao, N.~A. Smith, and
  Y.~Choi, ``Robust navigation with language pretraining and stochastic
  sampling,'' in \emph{Proceedings of the 2019 Conference on Empirical Methods
  in Natural Language Processing and the 9th International Joint Conference on
  Natural Language Processing (EMNLP-IJCNLP)}, 2019, pp. 1494--1499.

\bibitem{HOZ}
S.~Zhang, X.~Song, Y.~Bai, W.~Li, Y.~Chu, and S.~Jiang, ``Hierarchical
  object-to-zone graph for object navigation,'' in \emph{2021 IEEE/CVF
  International Conference on Computer Vision (ICCV)}, 2021, pp.
  15\,110--15\,120.

\bibitem{OA}
Y.~Qi, Z.~Pan, S.~Zhang, A.~van~den Hengel, and Q.~Wu, ``Object-and-action
  aware model for visual language navigation,'' in \emph{European Conference on
  Computer Vision}.\hskip 1em plus 0.5em minus 0.4em\relax Springer, 2020, pp.
  303--317.

\bibitem{VLMaps}
C.~Huang, O.~Mees, A.~Zeng, and W.~Burgard, ``Visual language maps for robot
  navigation,'' in \emph{2023 IEEE International Conference on Robotics and
  Automation (ICRA)}, 2023, pp. 10\,608--10\,615.

\bibitem{reduction}
S.~Ross, G.~Gordon, and D.~Bagnell, ``A reduction of imitation learning and
  structured prediction to no-regret online learning,'' in \emph{Proceedings of
  the fourteenth international conference on artificial intelligence and
  statistics}.\hskip 1em plus 0.5em minus 0.4em\relax JMLR Workshop and
  Conference Proceedings, 2011, pp. 627--635.

\bibitem{SMNA}
C.~Ma, J.~Lu, Z.~Wu, G.~AlRegib, Z.~Kira, R.~Socher, and C.~Xiong,
  ``Self-monitoring navigation agent via auxiliary progress estimation,'' in
  \emph{7th International Conference on Learning Representations, {ICLR} 2019,
  New Orleans, LA, USA, May 6-9, 2019}.\hskip 1em plus 0.5em minus 0.4em\relax
  OpenReview.net, 2019.

\bibitem{BUTD}
P.~Anderson, X.~He, C.~Buehler, D.~Teney, M.~Johnson, S.~Gould, and L.~Zhang,
  ``Bottom-up and top-down attention for image captioning and visual question
  answering,'' in \emph{2018 IEEE/CVF Conference on Computer Vision and Pattern
  Recognition}, 2018, pp. 6077--6086.

\bibitem{MP3D}
A.~Chang, A.~Dai, T.~Funkhouser, M.~Halber, M.~Niebner, M.~Savva, S.~Song,
  A.~Zeng, and Y.~Zhang, ``Matterport3d: Learning from rgb-d data in indoor
  environments,'' in \emph{2017 International Conference on 3D Vision
  (3DV)}.\hskip 1em plus 0.5em minus 0.4em\relax IEEE, 2017, pp. 667--676.

\bibitem{BERT}
J.~Devlin, M.-W. Chang, K.~Lee, and K.~Toutanova, ``{BERT}: Pre-training of
  deep bidirectional transformers for language understanding,'' in
  \emph{Proceedings of the 2019 Conference of the North {A}merican Chapter of
  the Association for Computational Linguistics: Human Language Technologies,
  Volume 1 (Long and Short Papers)}.\hskip 1em plus 0.5em minus 0.4em\relax
  Minneapolis, Minnesota: Association for Computational Linguistics, Jun. 2019,
  pp. 4171--4186.

\bibitem{LXMERT}
H.~H. Tan and M.~Bansal, ``Lxmert: Learning cross-modality encoder
  representations from transformers,'' in \emph{Conference on Empirical Methods
  in Natural Language Processing}, 2019.

\bibitem{pretrain1}
J.~Krantz, E.~Wijmans, A.~Majumdar, D.~Batra, and S.~Lee, ``Beyond the
  nav-graph: Vision-and-language navigation in continuous environments,'' in
  \emph{Computer Vision -- ECCV 2020}, A.~Vedaldi, H.~Bischof, T.~Brox, and
  J.-M. Frahm, Eds.\hskip 1em plus 0.5em minus 0.4em\relax Cham: Springer
  International Publishing, 2020, pp. 104--120.

\bibitem{pretrain2}
A.~Pashevich, C.~Schmid, and C.~Sun, ``Episodic transformer for
  vision-and-language navigation,'' \emph{2021 IEEE/CVF International
  Conference on Computer Vision (ICCV)}, pp. 15\,922--15\,932, 2021.

\bibitem{MRC}
J.~Lu, D.~Batra, D.~Parikh, and S.~Lee, ``Vilbert: Pretraining task-agnostic
  visiolinguistic representations for vision-and-language tasks,''
  \emph{Advances in neural information processing systems}, vol.~32, 2019.

\bibitem{ViT}
A.~Kolesnikov, A.~Dosovitskiy, D.~Weissenborn, G.~Heigold, J.~Uszkoreit,
  L.~Beyer, M.~Minderer, M.~Dehghani, N.~Houlsby, S.~Gelly, T.~Unterthiner, and
  X.~Zhai, ``An image is worth 16x16 words: Transformers for image recognition
  at scale,'' in \emph{Proceedings of the International Conference on Learning
  Representations (ICLR)}, 2021.

\bibitem{SAP}
S.~Chen, P.-L. Guhur, C.~Schmid, and I.~Laptev, ``History aware multimodal
  transformer for vision-and-language navigation,'' in \emph{NeurIPS}, 2021.

\bibitem{OG}
X.~Lin, G.~Li, and Y.~Yu, ``Scene-intuitive agent for remote embodied visual
  grounding,'' in \emph{2021 IEEE/CVF Conference on Computer Vision and Pattern
  Recognition (CVPR)}, 2021, pp. 7032--7041.

\bibitem{BEVBert}
D.~An, Y.~Qi, Y.~Li, Y.~Huang, L.~Wang, T.~Tan, and J.~Shao, ``Bevbert:
  Multimodal map pre-training for language-guided navigation,''
  \emph{Proceedings of the IEEE/CVF International Conference on Computer
  Vision}, 2023.

\bibitem{GBE}
M.~Li, Z.~Wang, T.~Tuytelaars, and M.-F. Moens, ``Layout-aware dreamer for
  embodied visual referring expression grounding,'' in \emph{Proceedings of the
  AAAI Conference on Artificial Intelligence}, vol.~37, 2023, pp. 1386--1395.

\end{thebibliography}
\end{document}